\documentclass[letterpaper]{article} 
\usepackage{aaai2026}  
\usepackage{times}  
\usepackage{helvet}  
\usepackage{courier}  
\usepackage[hyphens]{url}  
\usepackage{graphicx} 
\usepackage{natbib}  
\usepackage{caption} 
\usepackage{amsmath} 
\usepackage{amssymb}
\usepackage{algorithm}
\usepackage{algorithmic}

\usepackage{newfloat}
\usepackage{listings}
\DeclareCaptionStyle{ruled}{labelfont=normalfont,labelsep=colon,strut=off} 
\floatstyle{ruled}
\newfloat{listing}{tb}{lst}{}
\floatname{listing}{Listing}
\title{MoToRec: Sparse-Regularized Multimodal Tokenization for Cold-Start Recommendation}
\author {
    Jialin Liu\textsuperscript{\rm 1},
    Zhaorui Zhang\textsuperscript{\rm 2}\thanks{Corresponding author.},
    Ray C.C. Cheung\textsuperscript{\rm 1}
}
\affiliations {
    \textsuperscript{\rm 1} City University of Hong Kong\\
    \textsuperscript{\rm 2} The Hong Kong Polytechnic University\\
    
    camilla.liu@my.cityu.edu.hk, zhaorui.zhang@polyu.edu.hk, r.cheung@cityu.edu.hk
}

\usepackage{bibentry}

\begin{document}

\maketitle

\begin{abstract}

Graph neural networks (GNNs) have revolutionized recommender systems by effectively modeling complex user-item interactions, yet data sparsity and the item cold-start problem significantly impair performance, particularly for new items with limited or no interaction history. While multimodal content offers a promising solution, existing methods result in suboptimal representations for new items due to noise and entanglement in sparse data. To address this, we transform multimodal recommendation into discrete semantic tokenization. We present Sparse-Regularized \textbf{M}ultimodal \textbf{To}kenization for Cold-Start \textbf{Rec}ommendation (\textbf{MoToRec}), a framework centered on a sparsely-regularized Residual Quantized Variational Autoencoder (RQ-VAE) that generates a compositional semantic code of discrete, interpretable tokens, promoting disentangled representations. MoToRec's architecture is enhanced by three synergistic components: (1) a sparsely-regularized RQ-VAE that promotes disentangled representations, (2) a novel adaptive rarity amplification that promotes prioritized learning for cold-start items, and (3) a hierarchical multi-source graph encoder for robust signal fusion with collaborative signals. Extensive experiments on three large-scale datasets demonstrate MoToRec's superiority over state-of-the-art methods in both overall and cold-start scenarios. Our work validates that discrete tokenization provides an effective and scalable alternative for mitigating the long-standing cold-start challenge.

\end{abstract}


\section{Introduction}
\label{sec:introduction}

Graph Neural Networks (GNNs) have become the cornerstone of modern recommender systems, achieving state-of-the-art performance by modeling the rich connectivity of user-item interaction graphs \cite{he2020lightgcn, wu2021self, mo2024min}. However, their success relies on dense historical data, exposing a critical vulnerability that manifests as a sharp performance decline in the face of data sparsity, particularly the persistent item cold-start problem \cite{schein2002methods, li2024mhhcr}. To mitigate this, multimodal information offers a promising solution \cite{lu2025dmmd4sr, cui2025diffusion, cui2025multi}. Early methods simply concatenated content features with ID embeddings \cite{he2016vbpr}. More recent GNN-based methods have grown in sophistication: for instance, MMGCN \cite{wei2019mmgcn} builds modality-specific interaction graphs, LATTICE \cite{zhang2021mining} introduces item-item semantic graphs to capture latent correlations, and state-of-the-art approaches like FREEDOM \cite{zhou2023tale} and BM3 \cite{zhou2023bootstrap} further employ self-supervised contrastive learning to bridge the modality gap.

\begin{figure}[t!]
\centering
\includegraphics[width=\columnwidth]{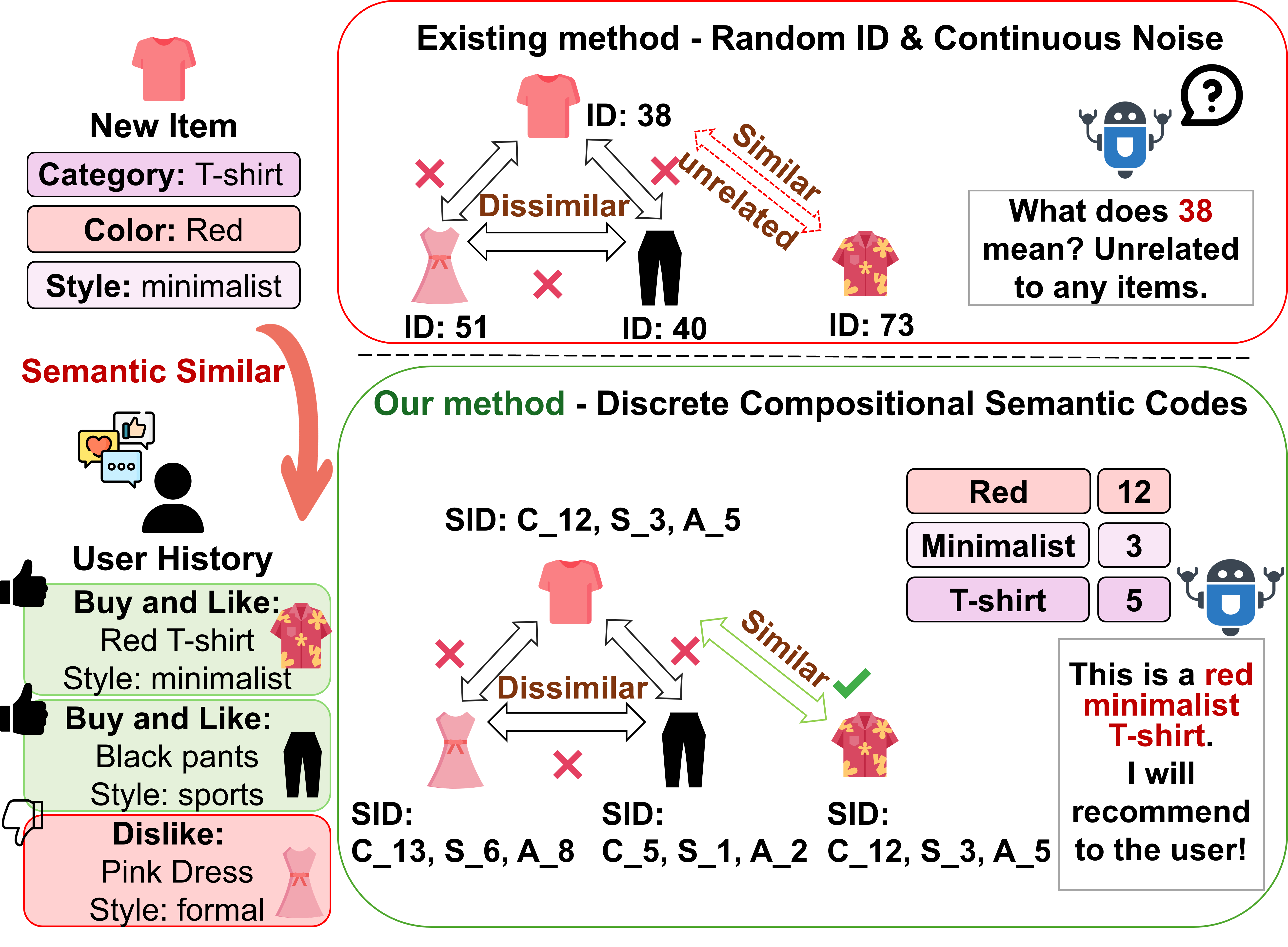}
\caption{From continuous alignment to discrete compositional codes. Top: Existing methods struggle with noisy alignment and uninformative IDs. Bottom: MoToRec generates robust and interpretable codes for effective cold-start recommendation.}
\label{fig:problem_illustration}
\end{figure}

Despite their architectural diversity and progress, these approaches are all fundamentally hampered by a shared challenge: the inherent ambiguity and noise in high-dimensional vector alignment. We term this phenomenon the \textbf{``semantic fog''}. It makes mapping a concept like a ``red T-shirt'' from pixel-based and text-based vectors into a single, coherent point in a high-dimensional space a noise-sensitive and unreliable task. This core issue persists even in the most recent advances leveraging powerful Large Language Models (LLMs) as feature extractors \cite{geng2022p5, bao2023tallrec, zhao2024meta, ye2024harnessing, liu2025coherency}. Aligning the noisy, continuous embeddings from these models often leads to suboptimal, out-of-distribution (OOD) representations, especially for cold-start items. As shown in Figure~\ref{fig:problem_illustration} above, this reliance on noisy alignment in a continuous space is a critical bottleneck limiting the effectiveness of existing methods.

To cut through this ``semantic fog'', we introduce a novel approach centered on discrete semantic tokenization. We present Sparse-Regularized \textbf{M}ultimodal \textbf{To}kenization for Cold-Start \textbf{Rec}ommendation (\textbf{MoToRec}), a framework that learns to convert raw multimodal features into a compositional semantic code. As shown in Figure~\ref{fig:problem_illustration} below, this code consists of a structured sequence of discrete tokens drawn from a learnable codebook, where each token represents a disentangled semantic concept (e.g., style: minimalist, color: red). 

In particular, the MoToRec framework operationalizes this vision through several synergistic components. At its core, we leverage a Residual Quantized Variational Autoencoder (RQ-VAE) \cite{lee2022} to generate the token sequence. To ensure these tokens are semantically meaningful, we devise a novel sparsity-inducing regularization to promote disentangled representations. Furthermore, to address the data imbalance inherent in recommendation, we introduce an adaptive rarity amplification mechanism to prioritize learning on less frequent items. Finally, a hierarchical multi-source graph encoder robustly fuses these newly created semantic codes with pure collaborative signals, aligning content-based understanding with user interaction patterns.

In summary, our main contributions are summarized as follows:
\begin{itemize}
    \item We propose a novel approach to multimodal recommendation, reframing it as a discrete semantic tokenization task to explicitly tackle the ``semantic fog'' and OOD issues prevalent in cold-start scenarios.
    \item We design MoToRec, an end-to-end architecture that synergistically integrates a sparsely-regularized RQ-VAE tokenizer, adaptive rarity amplification, and multi-source graph encoding for effective and robust signal fusion.
    \item We perform comprehensive experiments on three large-scale datasets to validate the effectiveness of our proposed approach, demonstrating significant improvements over state-of-the-art methods, particularly in cold-start scenarios.
\end{itemize}

\section{Related Work}

\subsection{Graph-based Multimodal Recommendation}
Graph Neural Networks (GNNs) \cite{kipf2017semi} have advanced recommender systems. Early graph-based multimodal methods directly inherited the message propagation mechanism to incorporate side information. For example, VBPR \cite{he2016vbpr} concatenated content features with ID embeddings, while MMGCN \cite{wei2019mmgcn} built modality-specific user-item graphs to learn distinct representations. LATTICE \cite{zhang2021mining}, for instance, constructs an item-item semantic graph to capture latent content correlations. More recently, contrastive learning has been widely adopted to enhance graph-based multimodal recommendations. State-of-the-art approaches like FREEDOM \cite{zhou2023tale} and BM3 \cite{zhou2023bootstrap} design sophisticated cross-modal contrastive tasks to better align representations from different modalities. However, by operating in a high-dimensional continuous space, these methods are susceptible to alignment noise, an issue exacerbated in sparse, cold-start scenarios. This limitation motivates our exploration of discrete representations.

\subsection{Vector Quantization in Recommendation}
Vector Quantization (VQ), with its roots in generative modeling \cite{van2017neural, zeghidour2021soundstream}, has been explored in recommender systems for its efficiency and noise-resilience. Early studies in this area primarily focused on embedding compression, where VQ is used to quantize large embedding tables to reduce memory footprint \cite{lian2020lightrec}. Later, VQ was adopted for generative sequence modeling. VQ-Rec \cite{hou2022vq}, for example, treats recommendation as a language modeling task over a discrete codebook of item prototypes, achieving strong performance in sequential recommendation. Unlike these works, which prioritize compression or sequence generation, our work is the first to leverage it to learn compositional, disentangled representations from multimodal content, specifically to address the item cold-start challenge.

\subsection{Cold-Start Recommendation}
The item cold-start problem, where new items have few or no interactions, is a long-standing challenge in recommender systems \cite{schein2002methods, bobadilla2013recommender}. Meta-learning offers one solution. Inspired by MAML \cite{finn2017model}, methods like MeLU \cite{lee2019melu} learn model initializations for rapid adaptation to new items with only a few examples. This ``few-shot'' approach, however, is ill-suited for the ``zero-shot'' scenario where no interactions exist. Another powerful direction leverages Large Language Models (LLMs), which have been explored as zero-shot feature extractors or even direct recommenders \cite{geng2022p5, bao2023tallrec, zhao2024meta, ye2024harnessing}. Despite their strong semantic capabilities, LLMs face two major hurdles in practice: (1) their continuous embeddings still suffer from the ``semantic fog'' alignment issue, and (2) their high computational cost challenges practical deployment. Our approach is motivated by this gap, aiming to distill knowledge into efficient, discrete codes for a scalable and robust zero-shot solution.

\section{Methodology}
\label{sec:methodology}

\begin{figure*}[t] 
    \centering
    \includegraphics[width=\textwidth]{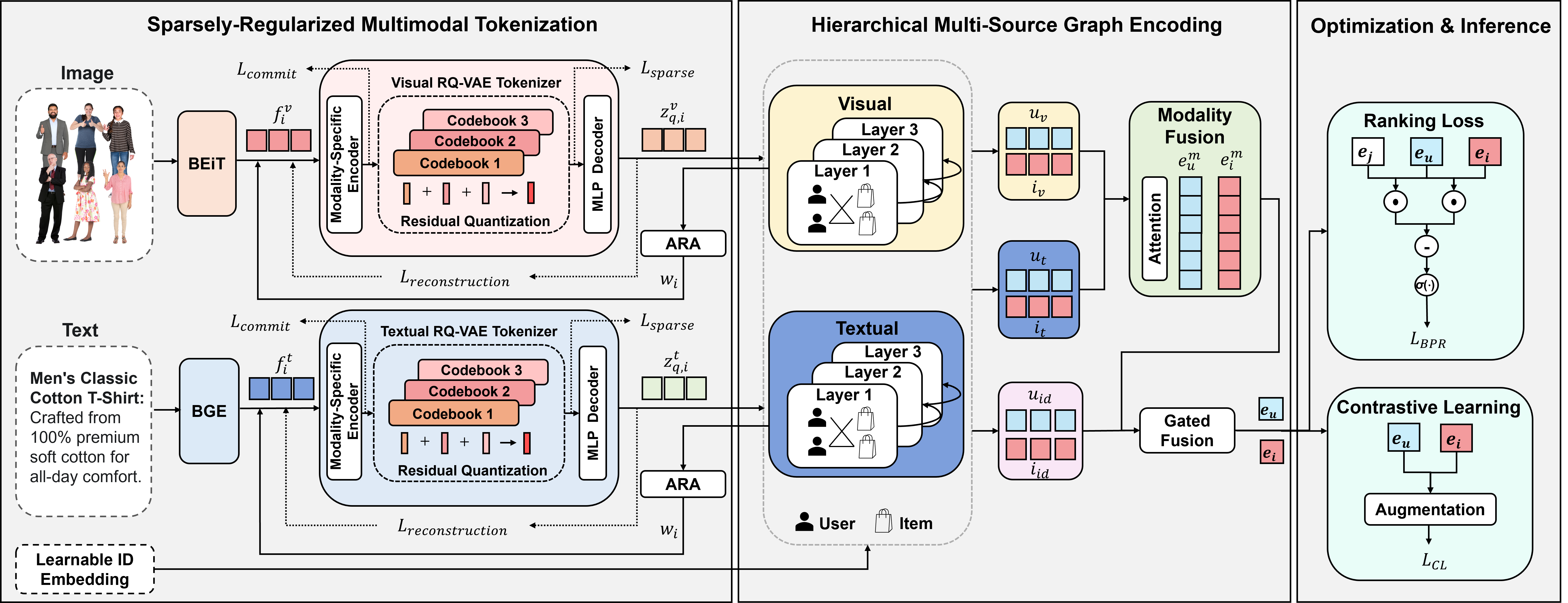} 
    \caption{The overall architecture of MoToRec. It consists of three main stages: (1)a sparsely-regularized multimodal tokenization module that converts raw features into discrete codes using RQ-VAEs; (2) a hierarchical multi-source graph encoding module to learn and fuse preferences; and (3) an optimization module. The optimization is guided by both ranking and self-supervised contrastive losses, and is made rarity-aware through a dynamic weighting scheme.}
    \label{fig:framework}
\end{figure*}

As illustrated in Figure~\ref{fig:framework}, MoToRec is a comprehensive framework designed to mitigate data sparsity and the item cold-start problem. It achieves this through three core components: an adaptive rarity amplification mechanism, a sparsely-regularized multimodal tokenizer, and a hierarchical multi-source graph encoder. We detail these components below.

\subsection{Problem Formulation}
Let \(\mathcal{U}\) denote the set of users and \(\mathcal{I}\) denote the set of items. The historical user-item interactions are represented by a sparse binary matrix \(\mathbf{R} \in \{0,1\}^{|\mathcal{U}| \times |\mathcal{I}|}\), where \(R_{ui}=1\) signifies an implicit feedback (e.g., a click or purchase) from user \(u\) to item \(i\). Each item \(i \in \mathcal{I}\) is endowed with rich multimodal content features: a visual feature vector \(\mathbf{f}_i^v \in \mathbb{R}^{d_v}\) extracted from a pre-trained vision transformer BEiT \cite{wang2023beit}, and a textual feature vector \(\mathbf{f}_i^t \in \mathbb{R}^{d_t}\) derived from a state-of-the-art sentence embedding model BGE \cite{su2023one}.

The primary objective is to learn a mapping function that produces expressive, low-dimensional embeddings for users, \(\mathbf{e}_u \in \mathbb{R}^{d}\), and items, \(\mathbf{e}_i \in \mathbb{R}^{d}\). These embeddings are then used to compute a relevance score via dot product \(\hat{y}_{ui} = \mathbf{e}_u^\top \mathbf{e}_i\), to rank items for each user. Our central challenge is to ensure the quality of \(\mathbf{e}_i\) for all items, especially those in the cold-start regime where interaction data is minimal. This disconnection is particularly acute for cold-start users, where abstract preferences (e.g., style or genre) must be inferred from content alone.

\subsection{Adaptive Rarity Amplification}
\label{sec:cold_start}
To directly combat the popularity bias inherent in recommendation datasets, which causes models to neglect rare items and underperform in representing items transitioning from \textit{cold} to \textit{warm}, we devise a dynamic, degree-aware weighting scheme. The goal of this component is to amplify the learning signal for less frequent items, ensuring the model pays sufficient attention to the very items that define the cold-start challenge.

First, we stratify items into \textit{cold} and \textit{warm} sets. The interaction degree of each item \(i\) is computed as \(d_i = \sum_{u \in \mathcal{U}} R_{ui}\). An item is designated as cold-start if its degree \(d_i\) falls below a domain-specific threshold \(\tau\):
\begin{equation}
    c_i = \mathbb{I}(d_i < \tau),
\end{equation}
where \(\mathbb{I}(\cdot)\) is the indicator function.

Subsequently, we formulate an item-specific weight, \(w_i\), designed to amplify the learning signal for items that are rare but not entirely devoid of interactions. The weight is defined by an inverse logarithmic relationship with the item's degree:
\begin{equation}
\label{eq:cold_weight}
w_i = 
\begin{cases} 
    \left( \log_2(d_i + 2) \right)^{-1} & \text{if } c_i = 1 \text{ and } d_i > 0, \\
    1.0 & \text{otherwise}.
\end{cases}
\end{equation}
This inverse logarithmic weighting \cite{schnabel2016recommendations, joachims2017unbiased} compresses degree ranges and stabilizes small values with a \(+2\) offset. The resulting item-specific weights \(w_i\) amplify the learning signal for less frequent items and are systematically integrated into the main learning objective.

\subsection{Sparsely-Regularized Multimodal Tokenization}
\label{sec:rqvae}
To overcome the ``semantic fog'' of continuous feature alignment and create robust semantic representations for cold-start items, we propose to transform raw multimodal features into a structured, discrete vocabulary of semantic tokens. This is implemented via a Residual Quantized Variational Autoencoder (RQ-VAE) \cite{lee2022}, which bridges the modality gap between continuous features and discrete collaborative IDs.

\paragraph{Residual Quantization.}
For each modality \(m \in \{v, t\}\), a modality-specific encoder \(E_m\) (an MLP) projects the raw feature \(\mathbf{f}_i^m\) into a latent space: \(\mathbf{z}_{e,i}^m = E_m(\mathbf{f}_i^m)\). This latent vector is then quantized by a cascade of \(N_q\) quantizers. At the first stage (\(k=1\)), the vector quantizer finds the closest prototype \(\mathbf{q}_i^{(1)}\) from a codebook \(\mathcal{C}_m^{(1)}\). The residual, \(\mathbf{r}_i^{(1)} = \mathbf{z}_{e,i}^m - \mathbf{q}_i^{(1)}\), is then passed to the second stage. This process repeats iteratively:
\begin{align}
    q^{(k)}_i &= \arg\min_{c \in C^{(k)}_m} \| r^{(k-1)}_i - c \|_2^2, \\
    r^{(k)}_i &= r^{(k-1)}_i - q^{(k)}_i,
\end{align}
The final quantized representation, \(\mathbf{z}_{q,i}^m = \sum_{k=1}^{N_q} \mathbf{q}_i^{(k)}\), is a composite vector that represents the item's multimodal characteristics as a combination of learned semantic primitives.

\paragraph{Sparsity-Inducing Regularization.}
A key challenge with learned codebooks is their susceptibility to the ``semantic fog'', producing entangled representations. To address this, we introduce a novel sparsity constraint on the codebook usage. This encourages the model to represent each item using a small, specialized subset of codebook vectors, making the resulting codes more explainable. We achieve this by imposing a KL-divergence penalty that drives the aggregate posterior distribution of codebook usage towards a sparse prior, specifically a Bernoulli distribution with a small mean \(\rho\). Let \(\hat{\rho}_j\) be the average activation probability of the \(j\)-th codeword across a mini-batch. The sparsity loss is:
\begin{equation}
\begin{split}
\label{eq:sparse_loss_fixed}
    \mathcal{L}_{\text{sparse}} = \sum_{j=1}^{K} \text{KL}(\rho \| \hat{\rho}_j) = \sum_{j=1}^{K} \Big( \rho \log \frac{\rho}{\hat{\rho}_j}\\
    + (1-\rho) \log \frac{1-\rho}{1-\hat{\rho}_j} \Big).
\end{split}
\end{equation}
This loss term encourages an efficient semantic code by penalizing high entropy in the average codebook usage. Theoretically, this KL penalty fosters disentangled representations by serving as a proxy for minimizing mutual information between codebook activations \cite{higgins2017beta}, driving the codebook usage towards a sparse prior. This process is analogous to nonlinear Independent Component Analysis in the discrete latent space, yielding compositional representations crucial for robust generalization to cold-start items.

\paragraph{Tokenizer Training Objective.}
The RQ-VAE for each modality is trained with a composite objective that includes a reconstruction term, a commitment term to stabilize codebook learning, and our novel sparsity term:
\begin{equation}
\label{eq:rq-vae-loss_fixed}
\begin{split}
\mathcal{L}_{\text{RQ-VAE}}^m = \underbrace{\|\mathbf{f}_i^m - D_m(\mathbf{z}_{q,i}^m)\|_2^2}_{\text{Reconstruction}} + \beta \underbrace{\|\mathbf{z}_{e,i}^m - \text{sg}(\mathbf{z}_{q,i}^m)\|_2^2}_{\text{Commitment}} \\
+ \gamma \underbrace{\mathcal{L}_{\text{sparse}}}_{\text{Sparsity}},
\end{split}
\end{equation}
where \(D_m\) is the decoder, \(\text{sg}(\cdot)\) is the stop-gradient operator, and \(\beta, \gamma\) are hyperparameters. This entire loss is weighted by \(w_i\) during global optimization, ensuring high-fidelity tokenization for cold-start items.

\subsection{Hierarchical Multi-Source Encoding and Fusion}
\label{sec:graph_encoding}
Having generated high-fidelity semantic codes from content, the subsequent critical step is to align these codes with users' collaborative preferences, thereby mitigating the OOD representation problem for cold-start items. To achieve this, we design a hierarchical graph encoding architecture to synthesize these diverse signals, built upon the efficient LightGCN \cite{he2020lightgcn}.

\paragraph{Intra-Modal Disentangled Propagation.}
Before fusing signals, it is critical to preserve the semantic purity of each information source. To this end, we maintain three parallel, disentangled propagation channels on the user-item graph \(\mathcal{G}_{ui}\), allowing us to learn modality-specific collaborative patterns without premature interference. The visual channel is initialized with tokenized visual embeddings \(\{\mathbf{z}_{q,i}^v\}_{i \in \mathcal{I}}\) to capture aesthetic preferences. The textual channel uses \(\{\mathbf{z}_{q,i}^t\}_{i \in \mathcal{I}}\) to learn from item attributes. Crucially, a separate collaborative channel, initialized with standard learnable ID embeddings, exclusively models pure collaborative signals, untainted by content. Within each channel, we apply the LightGCN propagation rule to refine embeddings over \(L\) layers:
\begin{equation}
    \mathbf{E}^{(l+1)} = (\mathbf{D}^{-1/2} \tilde{\mathbf{A}} \mathbf{D}^{-1/2}) \mathbf{E}^{(l)},
\end{equation}
where \(\tilde{\mathbf{A}}\) is the adjacency matrix of \(\mathcal{G}_{ui}\) with self-loops and \(\mathbf{D}\) is the diagonal degree matrix. Aggregating layer embeddings yields three specialized representations: \((\mathbf{u}_v, \mathbf{i}_v)\), \((\mathbf{u}_t, \mathbf{i}_t)\), and \((\mathbf{u}_{id}, \mathbf{i}_{id})\).

\paragraph{Cross-Source Fusion and Enhancement.}
To form the final representations, we fuse the specialized representations learned in the previous stage, designed to integrate content-based features with pure collaborative signals. We employ a hybrid fusion strategy:
\begin{equation}
\mathbf{e}_i^m = \alpha \cdot \text{CONCAT}(\mathbf{i}_v, \mathbf{i}_t) + (1-\alpha) \cdot \text{Attention}(\mathbf{i}_v, \mathbf{i}_t),
\end{equation}
where the hyperparameter \(\alpha\) balances static feature preservation with dynamic, context-aware re-weighting. This unified multimodal content embedding \(\mathbf{e}_i^m\) is then integrated with the collaborative embedding \(\mathbf{i}_{id}\) using a gated residual connection. The final user embedding \(\mathbf{e}_{u,\text{final}}\) is derived similarly from the user-side representations. This process yields the final predictive embeddings, \(\mathbf{e}_{u,\text{final}}\) and \(\mathbf{e}_{i,\text{final}}\).

\subsection{Model Optimization}
\label{sec:optimization}
The entire MoToRec framework is trained end-to-end by minimizing a composite objective function, prioritizing representation quality and high-fidelity tokenization for cold-start items. The final prediction score for a user-item pair is the dot product of their final embeddings: \(\hat{y}_{ui} = (\mathbf{e}_{u,\text{final}})^\top \mathbf{e}_{i,\text{final}}\).

The primary objective is the Bayesian Personalized Ranking (BPR) loss, which optimizes for the relative ranking of items over observed interactions \cite{rendle2009bpr}:
\begin{equation}
    \mathcal{L}_{\text{BPR}} = -\sum_{(u,i,j) \in \mathcal{D}} \ln \sigma(\hat{y}_{ui} - \hat{y}_{uj}),
\end{equation}
where \(\mathcal{D}\) is the set of training triplets where user \(u\) interacted with item \(i\) (positive) but not item \(j\) (negative).

To improve embedding quality, we incorporate the InfoNCE contrastive loss \cite{oord2018representation}, which pulls together augmented positive views of the same node while pushing apart negative samples:
\begin{equation}
    \mathcal{L}_{\text{CL}} = -\sum_{k \in \mathcal{B}} \log \frac{\exp(\text{sim}(\mathbf{e}_k^{(1)}, \mathbf{e}_k^{(2)})/\tau_{cl})}{\sum_{j} \exp(\text{sim}(\mathbf{e}_k^{(1)}, \mathbf{e}_j^{(2)})/\tau_{cl})},
\end{equation}
where \(\mathcal{B}\) is the mini-batch, and \(\text{sim}(\cdot, \cdot)\) denotes cosine similarity.

The final loss function integrates the ranking objective, contrastive loss, weighted multimodal tokenization loss, and a standard L2 regularization on all model parameters \(\Theta\):
\begin{equation}
\label{eq:total_loss_fixed}
\begin{split}
\mathcal{L} = \mathcal{L}_{\text{BPR}} + \lambda_{cl}\mathcal{L}_{\text{CL}} + \lambda_{rq} \sum_{m \in \{v,t\}} \frac{1}{|\mathcal{B}|} \sum_{i \in \mathcal{B}} w_i \cdot \mathcal{L}_{\text{RQ-VAE},i}^m \\
+ \lambda_{reg}\|\Theta\|_2^2,
\end{split}
\end{equation}
where \(\lambda_{cl}, \lambda_{rq}, \lambda_{reg}\) are hyperparameters that balance the different components. The rarity-amplification weight \(w_i\) ensures prioritized optimization for accurately tokenizing and representing cold-start items, which is central to our approach.

\section{Experiments}
\label{sec:experiments}


\subsection{Experimental Setup}

\paragraph{Datasets.}
We evaluate our model on three public Amazon review datasets\cite{mcauley2015}: Baby, Sports, and Clothing. To ensure a fair comparison, we follow the same data processing and filtering settings as in prior works that use these common benchmark datasets \cite{zhou2023tale, zhou2023bootstrap}. We utilize their rich user-item interaction, visual, and textual features. Detailed statistics are presented in Table~\ref{tab:datasets}.

\begin{table}[t!] 
\centering
\begin{tabular}{lrrrr} 
\hline
\hline
\textbf{Dataset} & \textbf{\#Users} & \textbf{\#Items} & \textbf{\#Inters.} & \textbf{Sparsity} \\
\hline
Baby & 19,445 & 7,050 & 160,792 & 99.88\% \\
Sports & 35,598 & 18,357 & 296,337 & 99.95\% \\
Clothing & 39,387 & 23,033 & 278,677 & 99.97\% \\
\hline
\hline
\end{tabular}
\caption{Statistics of the experimental datasets. The high sparsity highlights the challenge.}
\label{tab:datasets}
\end{table}

\paragraph{Baselines.}
We compare MoToRec against a comprehensive suite of baseline models, which can be categorized into two groups:
\begin{itemize}
    \item \textbf{Traditional}: MF-BPR \cite{rendle2009bpr}, LightGCN \cite{he2020lightgcn}, SimGCL \cite{yu2022graph}, LayerGCN \cite{zhou2023layer-refined}.
    \item \textbf{Multimodal}: VBPR \cite{he2016vbpr}, MMGCN \cite{wei2019mmgcn}, DualGNN \cite{wang2021dualgnn}, SLMRec \cite{tao2022self}, LATTICE \cite{zhang2021mining}, FREEDOM \cite{zhou2023tale}, BM3 \cite{zhou2023bootstrap}, LGMRec \cite{guo2024lgmrec}, LPIC \cite{liu2025lpic}.
\end{itemize}

\paragraph{Evaluation Protocol.}
We use an 8:1:1 train/validation/test split for all interactions. The test set is then divided into overall and cold-start groups. Following common practice \cite{schein2002methods, zhou2023tale}, the cold-start group contains test items with fewer than 10 interactions in the training set. For performance evaluation, we adopt two widely-used ranking metrics: Recall@N (R@N) and NDCG@N (N@N) \cite{he2015trirank}. We report results for \(N \in \{10, 20\}\), averaged over all test users.

\paragraph{Implementation Details.}

All models are implemented in the MMRec framework \cite{zhou2023mmrec} with 64-dimensional embeddings and the Adam optimizer. For MoToRec, we conduct a grid search over key hyperparameters. For the sparse RQ-VAE module, we tune the number of quantizers $N_q \in \{4, 6, 8\}$ and codebook size $K \in \{256, 512, 1024\}$. The learning rate is searched in $\{10^{-3}, 5 \times 10^{-4}, 10^{-4}\}$. We optimize regularization weights: sparsity coefficient $\gamma \in \{0.01, 0.05, 0.1, 0.2\}$, RQ-VAE loss weight $\lambda_{rq} \in \{0.1, 0.5, 1.0, 2.0\}$, and contrastive loss weight $\lambda_{cl} \in \{0.01, 0.05, 0.1\}$. The number of GCN layers is fixed at $L=2$, and the cold-start threshold is set to $\tau=10$, a standard value for defining sparse items \cite{schein2002methods}. Early stopping is applied with a patience of 20 epochs, monitoring R@20 on the validation set.




\subsection{Performance Comparison}
\begin{figure}[t!]
\centering
\includegraphics[width=\columnwidth]{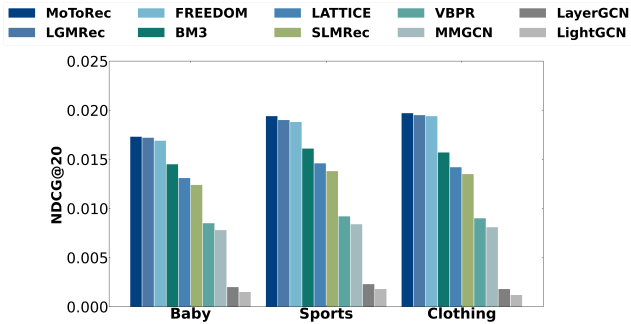}
\caption{Performance comparison on the cold-start item set.}
\label{fig:performance_analysis}
\end{figure}

The main performance comparison is summarized in Table~\ref{tab:main_results}, with key findings as follows:
(1) MoToRec's superiority over all baselines. MoToRec consistently outperforms all baselines, with gains up to 88\% over ID-only models (MF-BPR, LightGCN) and 11.57\% over state-of-the-art multimodal methods (LGMRec, LPIC). This highlights the value of multimodal content and, more critically, the superiority of our discrete representation.
(2) Effectiveness in mitigating the cold-start problem. MoToRec's advantage is most pronounced in the critical item cold-start scenario, as shown in Figure~\ref{fig:performance_analysis}. It achieves a remarkable uplift of up to 12.58\% in N@20 on items with the fewest interactions. This provides strong evidence that our discrete tokenization enables superior generalization by representing novel items as a composition of known concepts.
(3) Impact of sparse-regularized tokenization. The performance gap over strong baselines like LGMRec suggests that transforming features into discrete codes via our sparse-regularized RQ-VAE is key. This approach mitigates the ``alignment haze'' inherent in continuous fusion, creating more robust modality-aware representations. The importance of our content modeling is further corroborated by our ablation study in Table~\ref{tab:ablation_combined}.

\begin{table*}[t!]
\centering
\small
\setlength{\tabcolsep}{2pt}

\begin{tabular}{l|cccc|cccc|cccc}
\hline 
\hline 

\multicolumn{1}{c|}{\textbf{Datasets}} & \multicolumn{4}{c|}{\textbf{Baby}} & \multicolumn{4}{c|}{\textbf{Sports}} & \multicolumn{4}{c}{\textbf{Clothing}} \\
\cline{2-13}

\textbf{Model} & R@10 & R@20 & N@10 & N@20 & R@10 & R@20 & N@10 & N@20 & R@10 & R@20 & N@10 & N@20 \\
\hline 

MF-BPR   & 0.0357 & 0.0575 & 0.0192 & 0.0249 & 0.0432 & 0.0653 & 0.0241 & 0.0298 & 0.0187 & 0.0279 & 0.0103 & 0.0126 \\
LightGCN & 0.0479 & 0.0754 & 0.0257 & 0.0328 & 0.0569 & 0.0864 & 0.0311 & 0.0387 & 0.0340 & 0.0526 & 0.0188 & 0.0236 \\
SimGCL   & 0.0513 & 0.0804 & 0.0273 & 0.0350 & 0.0601 & 0.0919 & 0.0327 & 0.0414 & 0.0356 & 0.0549 & 0.0195 & 0.0244 \\
LayerGCN & 0.0529 & 0.0820 & 0.0281 & 0.0355 & 0.0594 & 0.0916 & 0.0323 & 0.0406 & 0.0371 & 0.0566 & 0.0200 & 0.0247 \\
\hline

Improv.  & \textbf{34.03\%} & \textbf{31.34\%} & \textbf{33.81\%} & \textbf{33.24\%} & \textbf{30.45\%} & \textbf{26.55\%} & \textbf{32.42\%} & \textbf{27.78\%} & \textbf{85.44\%} & \textbf{79.15\%} & \textbf{88.00\%} & \textbf{84.62\%} \\
\hline

VBPR     & 0.0423 & 0.0663 & 0.0223 & 0.0284 & 0.0558 & 0.0856 & 0.0307 & 0.0384 & 0.0281 & 0.0415 & 0.0158 & 0.0192 \\
MMGCN    & 0.0378 & 0.0615 & 0.0200 & 0.0261 & 0.0370 & 0.0605 & 0.0193 & 0.0254 & 0.0218 & 0.0345 & 0.0110 & 0.0142 \\
DualGNN  & 0.0448 & 0.0716 & 0.0240 & 0.0309 & 0.0568 & 0.0859 & 0.0310 & 0.0385 & 0.0454 & 0.0683 & 0.0241 & 0.0299 \\
SLMRec   & 0.0529 & 0.0775 & 0.0290 & 0.0353 & 0.0663 & 0.0990 & 0.0365 & 0.0450 & 0.0452 & 0.0675 & 0.0247 & 0.0303 \\
LATTICE  & 0.0547 & 0.0850 & 0.0292 & 0.0370 & 0.0620 & 0.0953 & 0.0335 & 0.0421 & 0.0492 & 0.0733 & 0.0268 & 0.0330 \\
FREEDOM  & 0.0627 & \underline{0.0992} & 0.0330 & 0.0424 & 0.0717 & 0.1089 & 0.0385 & 0.0481 & \underline{0.0628} & \underline{0.0941} & \underline{0.0341} & \underline{0.0420} \\
BM3      & 0.0564 & 0.0883 & 0.0301 & 0.0383 & 0.0656 & 0.0980 & 0.0355 & 0.0438 & 0.0422 & 0.0621 & 0.0231 & 0.0281 \\
LGMRec   & \underline{0.0639} & 0.0989 & \underline{0.0337} & \underline{0.0430} & 0.0719 & 0.1068 & 0.0387 & 0.0477 & 0.0555 & 0.0828 & 0.0302 & 0.0371 \\
LPIC     & 0.0634 & 0.0977 & \underline{0.0337} & 0.0422 & \underline{0.0737} & \underline{0.1113} & \underline{0.0398} & \underline{0.0485} & 0.0627 & 0.0928 & 0.0338 & 0.0405 \\
\hline 

MoToRec  & \textbf{0.0709} & \textbf{0.1077} 
         & \textbf{0.0376} & \textbf{0.0473}
         & \textbf{0.0784} & \textbf{0.1163} 
         & \textbf{0.0433} & \textbf{0.0529}
         & \textbf{0.0688} & \textbf{0.1014} 
         & \textbf{0.0376} & \textbf{0.0456} \\

Improv.  & \textbf{10.95\%} & \textbf{8.57\%} & \textbf{11.57\%} & \textbf{10.00\%} & \textbf{6.38\%} & \textbf{4.49\%} & \textbf{8.79\%} & \textbf{9.07\%} & \textbf{9.55\%} & \textbf{7.76\%} & \textbf{10.26\%} & \textbf{8.57\%} \\
\hline 
\hline
\end{tabular}
\caption{Overall performance comparison on three datasets. Best and second-best (best baseline) results are in \textbf{bold} and \underline{underlined}, respectively. The ``Improv.'' row shows the improvement of MoToRec over the best-performing baseline in each group. The t-tests validate the significance of performance improvements with p-value $\leq$ 0.05.}
\label{tab:main_results}
\end{table*}

\subsection{Ablation Study}
\label{sec:ablation}

To validate the contribution of each key component, we conduct a comprehensive ablation study with results detailed in Table~\ref{tab:ablation_combined}. The results clearly show that the full MoToRec model substantially outperforms all variants, confirming our design choices contribute synergistically. The most severe performance degradation stems from removing the entire RQ-VAE (\textit{w/o RQ-VAE}), providing strong evidence for our core thesis that discrete semantic tokenization is superior to continuous feature mapping, especially in cold-start scenarios. The components designed to combat data sparsity, Adaptive Rarity Amplification (\textit{w/o ARA}) and \textit{Sparsity} regularization, are also proven critical, as their removal significantly harms cold-start performance by failing to amplify rare signals and learn disentangled tokens. Finally, the consistent decline after removing the Contrastive Loss (\textit{w/o CL}), Hybrid Fusion (\textit{w/o HF}), and Homogeneous Graph Enhancement (\textit{w/o HGE}) verifies their respective importance in creating a robust embedding space, synthesizing multimodal signals, and capturing higher-order graph structures.

\begin{table*}[t!]
\centering
\small
\setlength{\tabcolsep}{3pt} 

\begin{tabular}{l|cc|cc|cc|cc|cc|cc}
\hline
\hline
\textbf{Datasets} & \multicolumn{4}{c|}{\textbf{Baby}} & \multicolumn{4}{c|}{\textbf{Sports}} & \multicolumn{4}{c}{\textbf{Clothing}} \\
\hline
\textbf{Scenario} & \multicolumn{2}{c|}{Overall} & \multicolumn{2}{c|}{Cold-start} & \multicolumn{2}{c|}{Overall} & \multicolumn{2}{c|}{Cold-start} & \multicolumn{2}{c|}{Overall} & \multicolumn{2}{c}{Cold-start} \\
\textbf{Metrics} & N@20 & R@20 & N@20 & R@20 & N@20 & R@20 & N@20 & R@20 & N@20 & R@20 & N@20 & R@20 \\
\hline
\textbf{MoToRec} & \textbf{0.0473} & \textbf{0.1077} & \textbf{0.0147} & \textbf{0.0347} & \textbf{0.0529} & \textbf{0.1163} & \textbf{0.0183} & \textbf{0.0452} & \textbf{0.0456} & \textbf{0.1014} & \textbf{0.0170} & \textbf{0.0420} \\
\hline
\textit{w/o RQ-VAE} & 0.0398 & 0.0915 & 0.0092 & 0.0229 & 0.0422 & 0.0930 & 0.0115 & 0.0304 & 0.0362 & 0.0816 & 0.0097 & 0.0281 \\
\textit{w/o ARA} & 0.0437 & 0.0977 & 0.0111 & 0.0281 & 0.0466 & 0.1027 & 0.0139 & 0.0367 & 0.0397 & 0.0894 & 0.0118 & 0.0342 \\
\textit{w/o Sparsity} & 0.0430 & 0.0972 & 0.0109 & 0.0277 & 0.0455 & 0.1003 & 0.0137 & 0.0362 & 0.0389 & 0.0876 & 0.0116 & 0.0336 \\
\textit{w/o CL} & 0.0455 & 0.1026 & 0.0118 & 0.0281 & 0.0515 & 0.1146 & 0.0153 & 0.0391 & 0.0438 & 0.0972 & 0.0129 & 0.0374 \\
\textit{w/o HF} & 0.0449 & 0.1028 & 0.0120 & 0.0303 & 0.0468 & 0.1042 & 0.0150 & 0.0396 & 0.0401 & 0.0899 & 0.0127 & 0.0368 \\
\textit{w/o HGE} & 0.0438 & 0.1002 & 0.0117 & 0.0295 & 0.0489 & 0.1106 & 0.0146 & 0.0384 & 0.0407 & 0.0917 & 0.0124 & 0.0357 \\
\hline
\hline
\end{tabular}
\caption{Ablation study of MoToRec on both overall and cold-start performance.}
\label{tab:ablation_combined}
\end{table*}



\subsection{Hyperparameter Study}
\label{sec:hyper_study}

\begin{figure}[t!]
\centering
\includegraphics[width=\columnwidth]{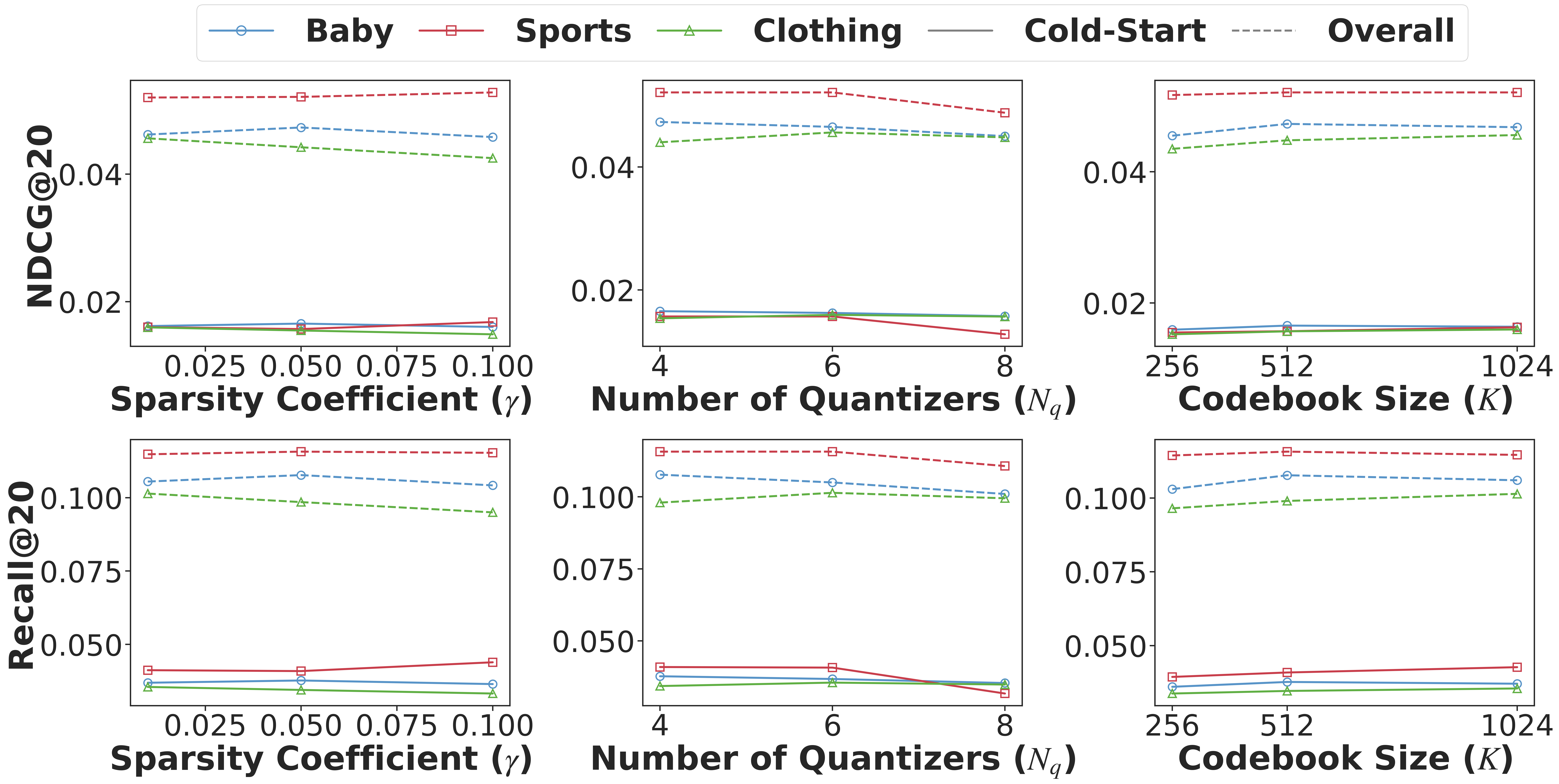}
\caption{Individual hyperparameter sensitivity analysis for N@20 and R@20 across all datasets for key parameters.}
\label{fig:hyperparam}
\end{figure}

\begin{figure}[t!]
\centering
\includegraphics[width=\columnwidth]{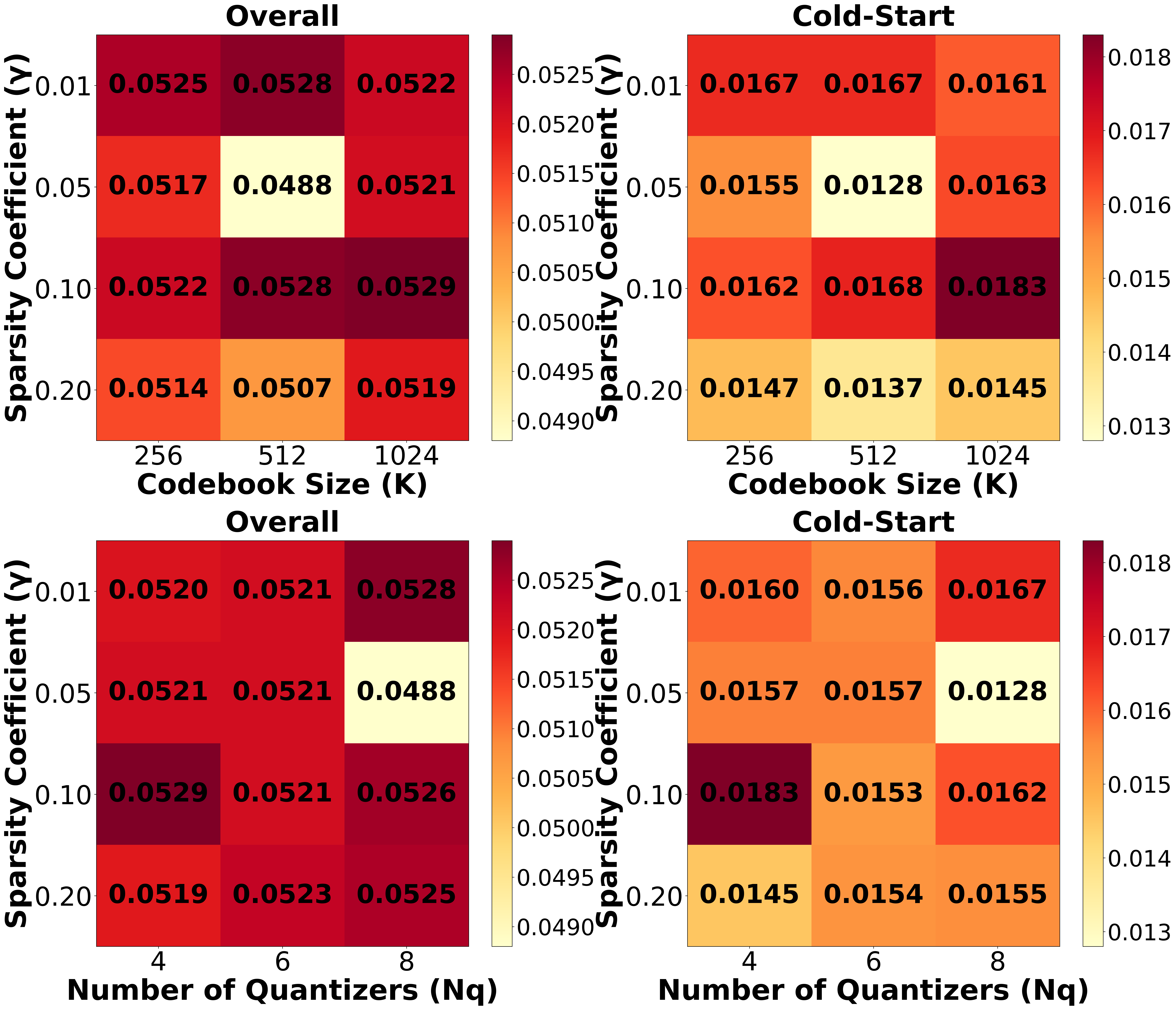}
\caption{Pairwise hyperparameter study on the Sports dataset (N@20).}
\label{fig:hyperparameter_study}
\end{figure}

We analyze hyperparameter sensitivity in Figures~\ref{fig:hyperparam} and \ref{fig:hyperparameter_study}. Results indicate that optimal configurations hinge on dataset characteristics: while sparse Baby dataset favors moderate sparsity ($\gamma=0.05$) and compact codebooks ($K=512$) for denoising, the visually rich Clothing dataset demands lower sparsity ($\gamma=0.01$) and larger capacity ($K=1024$) to preserve fine-grained semantics. Furthermore, pairwise interactions on Sports reveal that unlike the robust overall performance, cold-start results are highly sensitive to sparsity deviations, underscoring the necessity of precise tokenizer calibration for new items.

\subsection{Qualitative Analysis}
\label{sec:visualization}
\begin{figure}[t]
\centering
\includegraphics[width=\columnwidth]{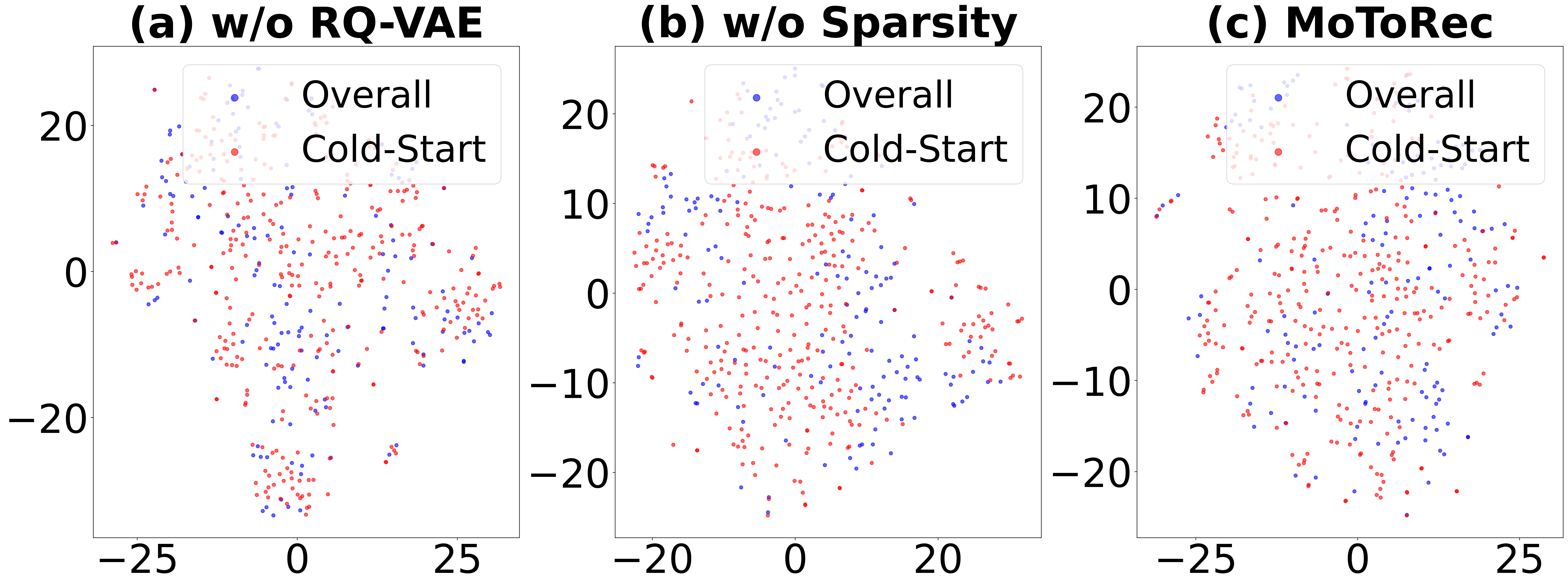}
\caption{t-SNE visualization of item embeddings.}
\label{fig:tsne}
\end{figure}

\paragraph{Embedding Space Visualization.}
To intuitively understand the quality of the learned representations, we visualize the item embedding space by sampling 500 items from the Sports dataset and projecting their embeddings into two dimensions using t-SNE~\cite{vanDerMaaten2008}, as shown in Figure~\ref{fig:tsne}. It confirms that the full MoToRec model (c) learns a significantly more organized semantic manifold compared to variants w/o RQ-VAE (a) or sparsity (b). Crucially, cold-start items (red) are no longer isolated outliers but are seamlessly integrated within this structure, proving our model's ability to position novel items near their semantic neighbors.

\paragraph{Case Study.}
We verify that our model's discrete codes learn human-interpretable concepts. On the Clothing dataset, individual codes consistently capture disentangled attributes; for example, code $<$c\_121$>$ reliably activates for the color `red', while $<$a\_34$>$ corresponds to the category `T-shirt'. This demonstrates our tokenizer learns a true semantic vocabulary. Critically, these codes are compositional: a cold-start item such as a `red minimalist T-shirt' activates a logical combination of these codes: [Color: Red] ($<$c\_121$>$), [Style: Minimalist] ($<$s\_5$>$), and [Category: T-shirt] ($<$a\_34$>$). This case study validates our model's ability to transform a single vector into a clear, compositional, and interpretable representation.

\subsection{Efficiency Study}

\begin{figure}[t!] 
\centering
\includegraphics[width=\columnwidth]{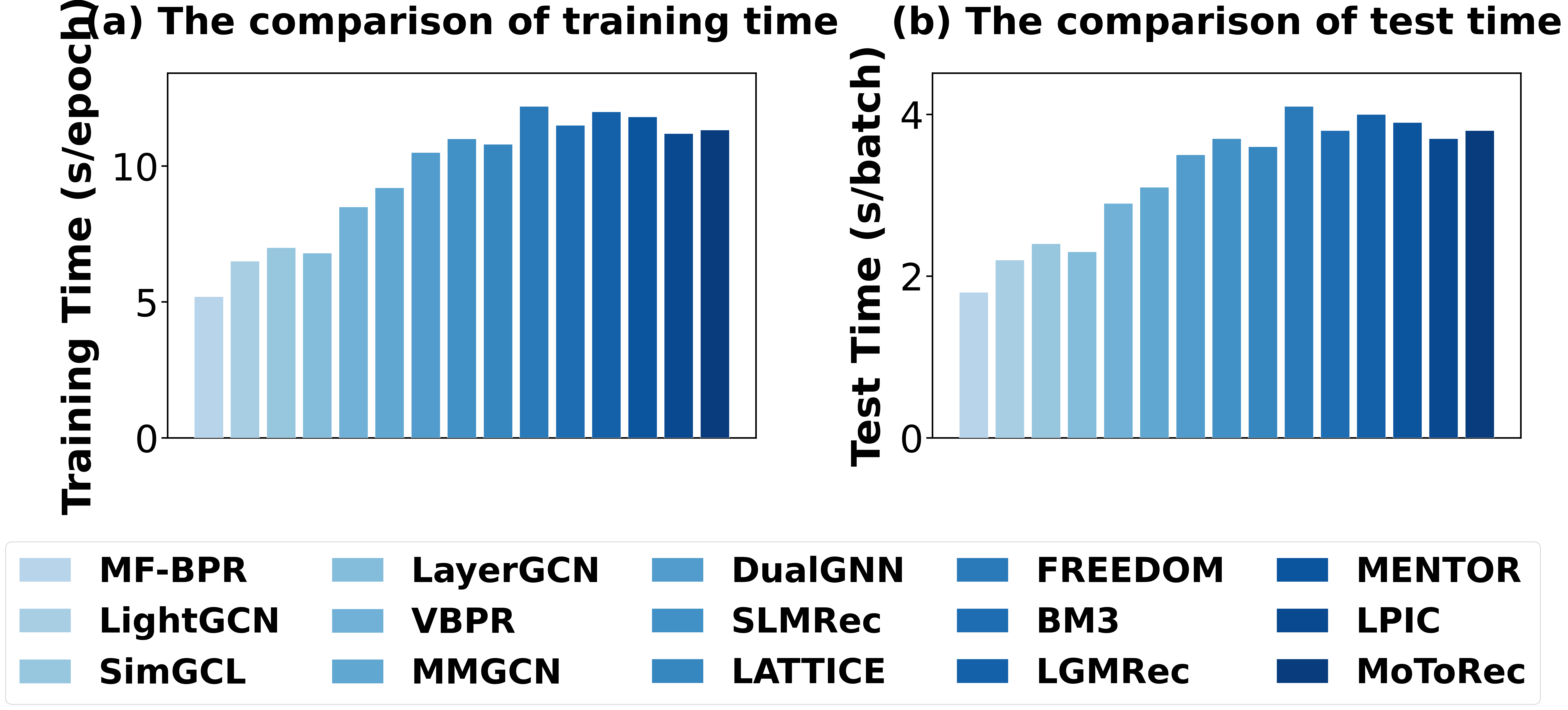}
\caption{Runtime comparison between MoToRec and baselines. (a) Training time per epoch. (b) Test time per batch.}
\label{fig:efficiency_comparison}
\end{figure}

To assess the practical deployability of MoToRec, we evaluate its computational efficiency. 
Figure~\ref{fig:efficiency_comparison} illustrates a detailed comparison of (a) training time per epoch and (b) test time per batch against a wide range of baseline models on the Sports dataset.
To validate MoToRec's practical viability, we benchmark its computational efficiency, with results presented in Figure~\ref{fig:efficiency_comparison}. MoToRec demonstrates a highly competitive training time of 11.33s per epoch, outperforming strong multimodal baselines such as FREEDOM at 12.2s and LGMRec at 12.0s. While this represents a modest $\sim$74\% overhead compared to the simpler ID-only LightGCN at 6.5s, this cost is solely attributed to our expressive tokenizer module; the graph propagation stage preserves the original efficiency of the LightGCN architecture. Critically, for inference, MoToRec remains highly practical with a test time of 3.8s per batch, on par with other high-performance models. This analysis confirms that MoToRec's architectural advancements do not impose a prohibitive computational cost, representing a favorable trade-off for its substantial accuracy gains, particularly on the item cold-start problem.

\section{Conclusion}

We propose MoToRec, which reframes recommendation as discrete semantic tokenization to learn robust multimodal representations from noisy raw data. Extensive experiments on three datasets demonstrate that MoToRec achieves state-of-the-art performance and substantial gains in cold-start scenarios. This validates discrete tokenization as a pivotal direction for future multimodal recommendations.

\bibliography{aaai2026}

\end{document}